\documentclass[10pt,letterpaper]{article}
\usepackage[top=0.85in,left=2.75in,footskip=0.75in]{geometry}

\usepackage{amsmath,amssymb}

\usepackage{changepage}
\usepackage{times}

\usepackage[utf8x]{inputenc}

\usepackage{textcomp,marvosym}

\usepackage{cite}

\usepackage{nameref,hyperref}

\usepackage[right]{lineno}

\usepackage{microtype}
\DisableLigatures[f]{encoding = *, family = * }

\usepackage[table]{xcolor}

\usepackage{array}

\newcolumntype{+}{!{\vrule width 2pt}}

\newlength\savedwidth

\raggedright
\usepackage[aboveskip=1pt,labelfont=bf,labelsep=period,justification=raggedright,singlelinecheck=off]{caption}

\makeatletter
\renewcommand{\@biblabel}[1]{\quad#1.}
\makeatother

\date{}

\usepackage{lastpage,fancyhdr,graphicx}
\usepackage{epstopdf}
\fancyheadoffset[L]{2.25in}
\fancyfootoffset[L]{2.25in}
\usepackage[square,sort,comma,numbers]{natbib}
\usepackage{verbatim}
\usepackage{framed}
\usepackage{subcaption}
\usepackage{framed}
\usepackage{color, colortbl}
\usepackage[colorinlistoftodos]{todonotes}
\usepackage{capt-of}
\newcommand\inlinetodo[1]{\textcolor{red}{#1}}
\definecolor{high}{rgb}{0.7,.85,.85}

\newcommand{\beginsupplement}{%
        \setcounter{table}{0}
        \renewcommand{\thetable}{S\arabic{table}}%
        \setcounter{figure}{0}
        \renewcommand{\thefigure}{S\arabic{figure}}%
}

\begin{document}
\vspace*{0.2in}

\begin{flushleft}
{\Large
\textbf\newline{Latent Human Traits in the Language of Social Media:\\ An Open-Vocabulary Approach} % Please use "title case" (capitalize all terms in the title except conjunctions, prepositions, and articles).
}
\newline
\\
Vivek Kulkarni\textsuperscript{1},
Margaret L. Kern\textsuperscript{2},
David Stillwell\textsuperscript{3},
Michal Kosinski\textsuperscript{4},
Sandra Matz\textsuperscript{5},
Lyle Ungar\textsuperscript{6},
Steven Skiena\textsuperscript{1},
H. Andrew Schwartz\textsuperscript{1}
\\
\bigskip
\textbf{1} Department of Computer Science, Stony Brook University
\\
\textbf{2} Melbourne Graduate School of Education, The University of Melbourne
\\
\textbf{3} Judge Business School, University of Cambridge
\\
\textbf{4} Graduate School of Business, Stanford University
\\
\textbf{5} Columbia Business School, Columbia University, New York, United States
\\
\textbf{6} Computer and Information Science, University of Pennsylvania
\\
\bigskip

% Insert additional author notes using the symbols described below. Insert symbol callouts after author names as necessary.
% 
% Remove or comment out the author notes below if they aren't used.
%
% Primary Equal Contribution Note
% \Yinyang These authors contributed equally to this work.

% Additional Equal Contribution Note
% Also use this double-dagger symbol for special authorship notes, such as senior authorship.
% \ddag These authors also contributed equally to this work.

% \textcurrency Current Address: Dept/Program/Center, Institution Name, City, State, Country % change symbol to "\textcurrency a" if more than one current address note
% \textcurrency b Insert second current address 
% \textcurrency c Insert third current address

% \dag Deceased

% Group/Consortium Author Note
% \textpilcrow Membership list can be found in the Acknowledgments section.

% Use the asterisk to denote corresponding authorship and provide email address in note below.
* has@cs.stonybrook.edu

\end{flushleft}
\section*{Abstract}
Over the past century, personality theory and research has successfully identified core sets of characteristics that consistently describe and explain fundamental differences in the way people think, feel and behave. 
Such characteristics were derived through theory, dictionary analyses, and survey research using explicit self-reports. 
The availability of social media data spanning millions of users now makes it possible to automatically derive 
%additional characteristics that distinguish individuals at large scale. 
characteristics from behavioral data --- language use --- at large scale.
Taking advantage of linguistic information available through Facebook, we study the process of inferring a new set of potential human traits based on \emph{unprompted} language use. 
We subject these new traits to a comprehensive set of evaluations and compare them with a popular five factor model of personality. 
We find that our language-based trait construct is often more \textit{generalizable} in that it often predicts non-questionnaire-based outcomes better than questionnaire-based traits (e.g. entities someone likes, income and intelligence quotient), while the factors remain nearly as \textit{stable} as traditional factors. 
Our approach suggests a value in new constructs of personality derived from \textit{everyday human language use}.
%\vivek{https://tex.stackexchange.com/questions/26146/figure-numbering-without-the-figure}

% Please keep the Author Summary between 150 and 200 words
% Use first person. PLOS ONE authors please skip this step. 
% Author Summary not valid for PLOS ONE submissions.   
%\section*{Author Summary}
%%linenumbers

% Use "Eq" instead of "Equation" for equation citations.
\section*{Introduction}
\emph{What are the fundamental characteristics that make a person uniquely him or herself?}
Psychology has long tried to answer this question by deriving the latent factors that distinguish people, are relatively stable across time and populations, and predict meaningful outcomes~\cite{allport1936trait,cattell1946description,mcadams1992five,john1999big}. While several different models of human personality exist, the most dominant model is the Big 5 or Five Factor Model, in which personality characteristics generally group into five factors: extraversion, agreeableness, conscientiousness, neuroticism/ emotional stability, and openness/ intellect \cite{costa1989neo,goldberg1993structure}.  %{also cite Goldberg, 1993%} 
%Goldberg, L. R. (1993). The structure of phenotypic personality traits. American Psychologist, 48, 26-34 .
The Big 5 is meant to summarize, at a broad level, characteristic behaviors that distinguish a person throughout different contexts of their daily life. 
It is typically assessed through a questionnaire, in which a person reflects on their typical thoughts and behaviors~\cite{costa1989neo}. 
Such questionnaires only indirectly capture actual behavior and also suffer from systematic response biases~\cite{goldberg1963model,javeline1999response}. 

The rise of big data offers opportunities to study everyday behavior at a scale never before possible. 
Each day, people reveal aspects of their lives through words expressed online through social media, such as Facebook and Twitter. 
Leveraging this linguistic information, we derive a trait model based on everyday linguistic behavior. 
Our approach analyzes the words and phrases of tens of thousands of users and their millions of messages to infer a set of traits. 

In line with trait theory~\cite{allport1927concepts}, we seek a small number of generalizable and stable traits that capture meaningful differences between people. 
We call these \textit{behavior-based linguistic traits (BLTs)}. 
Our method does not rely on any hand-crafted lexica or questionnaires and it scales well to leverage the large amount of data available on social media. 
While some have leveraged social media and open-vocabulary techniques to assess \textit{existing} trait models --- e.g., Big 5 \cite{costa1989neo} of bloggers~ \cite{iacobelli2011large} or Facebook users~\cite{schwartz2013personality}, the dark triad \cite{furnham2013dark} using Facebook~\cite{darktriad2016cikm} --- to the best of our knowledge, none have attempted to \textit{infer} the latent traits themselves. 

We evaluate BLTs along two criteria: 
\begin{itemize}
\item \textbf{Generalizability}: The factors need to be generalizable across a large variety of predictive tasks.
\item \textbf{Stability}: The factors should be relatively stable over time and populations. The factor scores of users over time should be correlated, at levels similar to cross-time and cross-population correlations of the Big 5. 
\end{itemize}

The overall goal of this study is to determine if it is feasible to derive \textit{generalizable} and \textit{stable} traits from the behavior of social media language use. 
We produce \textit{BLTs} using methods of matrix factorization. Then, we evaluate the extent to which the \textit{BLTs} meet this criteria by considering their predictive validity, test-retest validity, dropout reliability, face validity, and ability to predict psychological variables (\textsc{Depression Scores}) and social-demographic variables (\textsc{IQ}, \textsc{Income} and \textsc{Likes}). 
%as well as tests for temporal validity and dropout-validity.
%(see Section \ref{sec:fac_eval})

\section*{Background}
Personality theory and research has a long history of finding characteristics, factors, and aspects that distinguish people from each other. Although personality has been used in a myriad of ways, much of the research can generally be classified into two purposes: (a) Identifying the major constructs and factors that consistently distinguish groups of individuals, and (b) Predictive models that either predict personality traits from other characteristics or use personality to predict other variables. We further discuss each of these themes below.

\subsection*{Modeling Personality}
Personality psychologists have long sought to identify fundamental characteristics that ideographically distinguish individuals yet also adequately cluster across groups of individuals to reveal consistent patterns of behavior. While there are multiple approaches for studying individual differences, the most dominant approaches are rooted in the lexical hypothesis, which suggests that key features of human personality will become a part of the language that we use to describe ourselves \cite{allport1936trait,goldberg1981language,john1999big}. The importance of language in psychology is underscored by \cite{tausczik2010psychological}: 
\begin{quotation}
Language is the most common and reliable way for people to translate their internal thoughts and emotions into a form that others can understand. Words and language, then, are the very stuff of psychology and communication.
\end{quotation}
Consequently, a long line of work in personality  psychology has sought to characterize individual differences based on words that people use. The Big 5 model arose by having people rate themselves across dictionary-based adjectives (e.g., I see myself as extraverted, enthusiastic; reserved, quiet), and then using factor analysis to group responses into a small number of factors. Early on, a lexicon of $18,000$ words that distinguish one person from another based on an English dictionary was proposed by \cite{allport1936trait}. Participants rated themselves on these characteristics, and numerous factor analyses revealed anywhere between three and 16 major factors. Other lexica and lists of adjective descriptors have also been developed and tested over the past century. Although there are some exceptions, across multiple questionnaires, cultures, and temporal periods, the Big 5 factors consistently appear \cite{goldberg1993structure}. %cite Goldberg 1993%.

As such, the dominant approach towards characterizing personality is based on hand-crafted lexica and dictionaries \cite{john1999big}. Psychologists have developed a variety of questionnaires that capture these personality traits \cite{costa1989neo,gosling2003very,john1999big}. People can easily reflect on their own personality, or others can provide an evaluation of that person’s personality using these questionnaires.

The five factors are hierarchical in nature, with the five factors underscored by aspects which are comprised of specific traits or facets, which reflect habitual behaviors, thoughts, emotions, and ways of responding to situations \cite{deyoung2007between}. %cite DeYoung 2007
% DeYoung, C. G., Quilty, L. C., & Peterson, J. B. (2007). Between facets and domains: 10 aspects of the big five. Journal of Personality and Social Psychology, 93, 880-896.
Personality questionnaires are meant to summarize everyday behavior, but represent a reflection based on self (or other) perceptions, rather than directly assessing everyday behavior as it occurs. Underlying characteristics better predict outcomes \cite{paunonen2000beyond}, but also are less consistent across individuals. As the Big 5 do successfully predict important life outcomes, they are useful in providing broad representations of relatively stable individual differences, at the expense of capturing behaviors that occur in everyday life. 

%pk I removed the bulleted list of drawbacks, as I think that works better in the text. But the compiler doesn't seem to like what I did, as it just keeps giving me errors since I did this. Sorry if I messed it up!

Other approaches to personality assessment exist. For instance, narrative approaches to personality (e.g., \cite{mcadams2006role}) %cite McAdams, 2006 % 
% McAdams, D. P. (2006). The role of narrative in personality psychology today. Narrative Inquiry, 16, 11-18.
successfully provide rich idiographic descriptions of a person within their everyday context. However, such approaches are time and resource intensive, and thus only occur at small scale, making it a challenge to find common nomothetic constructs.

An initial attempt was made to use an open-ended approach to learn personality traits by extracting common themes from self-narrative texts \cite{chung2008revealing}. Using a dataset of $1,165$ open-ended self-descriptive narratives, a factor analysis was performed on the most frequently used adjectives to reveal latent factors. Latent factors were shown to correlate moderately with the Big 5 factors and suggested psychologically meaningful dimensions. The current study extends this work at much larger scale.

\subsection*{Predictive Models of Personality}
Personality is interesting in part because it is predictive of meaningful life outcomes, including education and job success, social relationships, physical and mental health, and longevity (e.g., \cite{friedman2014personality,roberts2007power}).
% cite Friedman & Kern, 2014; Roberts et al., 2007
% Friedman, H. S., & Kern, M. L. (2014). Personality, well-being, and health. Annual Review of Psychology, 65, 719-742.
% Roberts, B. W., Kuncel, N. R., Shiner, R., Caspi, A., & Goldberg, L. R. (2007). The power of personality: The comparative validity of personality traits, socioeconomic status, and cognitive ability for predicting important life outcomes. Perspectives on Psychological Science, 2, 313-345.
Personality can also be predicted by other biological, social, psychological, and behavioral measures. 

With recent advances in computational social science, researchers have sought to predict  personality, as well as other psychosocial and demographic differences, amongst users  \cite{argamon2007mining,holtgraves2011text,sumner2011determining,golbeck2011predicting,iacobelli2011large, sumner2012predicting,plank2015personality,park2015automatic,liu2016language,liu2016analyzing}. In this paper, we focus on work related to the description and prediction of personality from language on social media.

Multiple studies have analyzed social media data, including Facebook posts, text messages, and Twitter tweets, finding correlates of the Big 5 and other individual characteristics with language \cite{holtgraves2011text,sumner2011determining,kern2014online}.
% also cite Kern et al., 2014
% Kern, M. L., Eichstaedt, J. C., Schwartz, H. A., Dziurzynski, L., Ungar, L. H., Stillwell, D. J., … & Seligman, M. E. P. (2014). The online social self: An open vocabulary approach to personality. Assessment, 21, 158-169.
Park et al. developed a language-based personality assessment based on Facebook language that predicted personality at similar levels to other methods of personality assessment \cite{park2015automatic}. Golbeck et al. proposed a method to predict personality of a user based on their posts on Twitter \cite{golbeck2011predicting}. Sumner et al. proposed a method to predict dark personality traits, based on Twitter language \cite{sumner2012predicting}. Plank and Hovy analyzed $1.2$ million Tweets and proposed a model to predict Myers-Briggs personality types \cite{plank2015personality}. 

While successful, studies increasingly suggest various complexities. Iacobelli et al. analyzed bloggers and found that the best performing model combined multiple linguistic features, including stemmed bigrams and common words \cite{iacobelli2011large}. Notably, they also highlighted the need for more refined and complex linguistic features. Recent works show that moving beyond words and incorporating complex features like distributed sentence representations improve personality prediction \cite{liu2016language,liu2016analyzing}. As a whole, studies suggest that language can predict personality, but simplistic models might fail to adequately capture the complexities of the human psyche.

While existing studies begin with personality models as the ground truth, the current study harnesses the power of social media data to examine traits that arise from the language itself.

\section*{Materials and Methods}
\label{sec:methods}
In this section, we describe the details of our dataset, our proposed method to learn latent factors and the design of experiments to evaluate the derived factors.

\subsection*{Datasets}
We use a dataset of $20,356,117$ Facebook status messages over $152,845$ distinct users obtained using the \textsc{MyPersonality} application \cite{kosinski2015facebook}. 
We filter out all users who posted less than $1000$ words overall, are over 65 years of age, or who claim that they are not from the US. Additionally, we filter out all messages that are not English.
% PK question – what’s the DLA toolkit? Need to cite this or indicate what this is

Among these users, $49,139$ have data on age, gender and their Big 5 personality scores (\textsc{Big5}), based on 20 personality items from the International Personality Item Pool (IPIP) \cite{goldberg2006international} which comprises our final data set.
% for IPIP, cite Goldberg, L. R., Johnson, J. A., Eber, H. W., Hogan, R., Ashton, M. C., Cloninger, C. R., & Gough, H. G. (2006). The international personality item pool and the future of public domain personality measures. Journal of Research in Personality, 40, 84–96.
A few sample messages are shown below:
\begin{framed}
\texttt{\textbullet \; goodbye to anybody i didn't get to chill with before i left\\
\indent
\textbullet \;  a relaxed mind mks u see things in a better shade whose existance were merely ignored \\
\indent
\textbullet \; scars heal , glory fades and all we're left with are the memories made
}
\end{framed}
About $62.8\%$ of the users in our dataset are female. The age distribution is skewed towards younger people with the median age of $22$ years and a mean age of $25.49$ years.  While we believe that our derived latent factors should capture various ages and genders, we also investigate residualizing out  demographic factors (age and gender). All messages are tokenized and stop-words (very frequent words like ``the'' or ``is'') are removed in the pre-processing stage. 

\subsection*{Factor Generation}
We considered several methods to learn our latent factors. Before describing our proposed method, we discuss a few alternative methods that we considered to learn the latent factors.
\subsubsection*{Alternative methods}
\begin{itemize}
\item \textbf{Latent Dirichlet Allocation (LDA)}: We first modeled the compiled linguistic information from each user as a document and latent factors as topics.
%PK to say each user is a document rubbed me wrong, but saying their information is a document makes sense
LDA \cite{blei2003latent} enables us to learn these factors (i.e., topics) that represent each person based on the corpus of the user’s text. Each user is then represented as a mixture over the learned factors. We used the \textsc{MALLET} \cite{mccallum2002mallet} toolkit to learn LDA factors, set $\alpha=5$, and enabled hyper-parameter optimization.  We optimize the hyper-parameters every 20 iterations with a burn-in of 10 iterations. 

However, the LDA factors were negatively correlated with each other. This can be explained by noting that the factor scores obtained by LDA for each user must sum to 1. This implies that LDA, a probabilistic model, is not well-suited for modeling latent factors, and that probabilistic models are not suitable for modeling traits.

\item \textbf{Singular Value Decomposition (SVD)}: 
We next considered SVD \cite{landauer2006latent}. We specified a user-set ($\mathcal{U}$) and a finite set of vocabulary terms corresponding to those users ($\mathcal{V}$). We constructed a term-user matrix $M$ and computed a low-rank approximation of $M$ using SVD.

While SVD does not have the drawbacks of LDA, we observed that a more general version of dimensionality reduction, ``factor analysis (FA)'', demonstrates better empirical performance on predictive tasks and motivates FA as our proposed method to capture traits. As such, our proposed methodology draws on FA, rather than LDA or SVD.  

\end{itemize}
\subsubsection*{Proposed Method: Factor Analysis (FA)}
Factor Analysis (FA) has long been a part of psychological research and the development of psychometric assessments. The method seeks to represent a set of variables as linear combinations of a small number of latent factors. Formally, given a matrix $M$, factor analysis seeks to learn latent factors $F$ and a loading matrix $L$ such that:
\begin{equation}
M = FL + E
\end{equation}
where $E$ represents an error matrix. While SVD seeks to learn factors that account for all of the variance, FA is more general and learns factors that account for the common variance but allows for some residual variance not explained by the latent factors. 

Typically in psychology, factors are estimated and then rotated to find the best fitting solution. We chose to use FA as an approach to identify latent user factors. We applied FA on the User-Term matrix and investigated various rotations of the loading matrix L to obtain potentially more interpretable factors. In our final models, we used a promax (oblique) rotation with equa-max rotation for all our experiments.

%pk – what type of FA? There are multiple types, with different assumptions and results. And what rotation did you settle on? Orthogonal or oblique? Are results of these rotations in an appendix somewhere? Most readers won’t care, but some will want all of this available.
%vivek We have results of those rotations (as word clouds) on Google Drive.
%pk I'm happy with how this is described now, so all good

Using factor analysis, we extracted 5 factors. We chose to extract 5 factors as a direct comparison to the Big5 model. For comparison, we also extracted 10 and 30 factors (aligning with 10 underlying aspects \cite{deyoung2007between} %cite DeYoung, 2007)
and 30 facets \cite{costa1989neo}; results of these models are available in the Supplemental Information section. Figure \ref{fig:wordclouds_corr} illustrates word clouds corresponding to the most and least correlated words for each factor (i.e., different language analysis; \cite{schwartz2013personality}). 
% cite Schwartz et al., 2013 Schwartz, H. A., Eichstaedt, J. C., Kern, M. L., Dziurzynski, L., Ramones, S. M., Agrawal, M., … & Ungar, L. H. (2013). Personality, gender, and age in the language of social media: The open-vocabulary approach. PLOS ONE, 8, e73791. 
The factors capture  emotion words such as \texttt{life, heart, happiness, love} (see \textsc{FA:F1(+)}) and non-emotion words such as \texttt{week-end, school, work, tomorrow, tonight} (see \textsc{FA:F1(-)}).  These observations suggest that our factors capture a variety of behavioral and psychological cues. % \vivek{Check where you mention demographic cues and highlight.}
% PK – you noted demographic cues. What points to demographics? That’s not clear to me. But rather it seems like it’s psychological & social ones.
% vivek Ignore the comment below. This is old writing. Verified no mention of demographics.
%PK - all good

\ifdefined\submit
\subsection*{Fig 1. Word clouds showing the most/least correlated words for each factor (with rotation).}
{\small Word clouds showing the most/least correlated words for each FA factor (with rotation) as obtained using Differential Language Analysis (\cite{schwartz2013personality}). %\cite Schwartz et al., 2013%
The larger the word, the more strongly it correlates with the factor. Color indicates frequency (grey = low use, blue = moderate use, red = frequent use)}.
\refstepcounter{section}
\label{fig:wordclouds_corr}
\else
\begin{figure*}[htb!]
\centering
        \includegraphics[width=\textwidth]{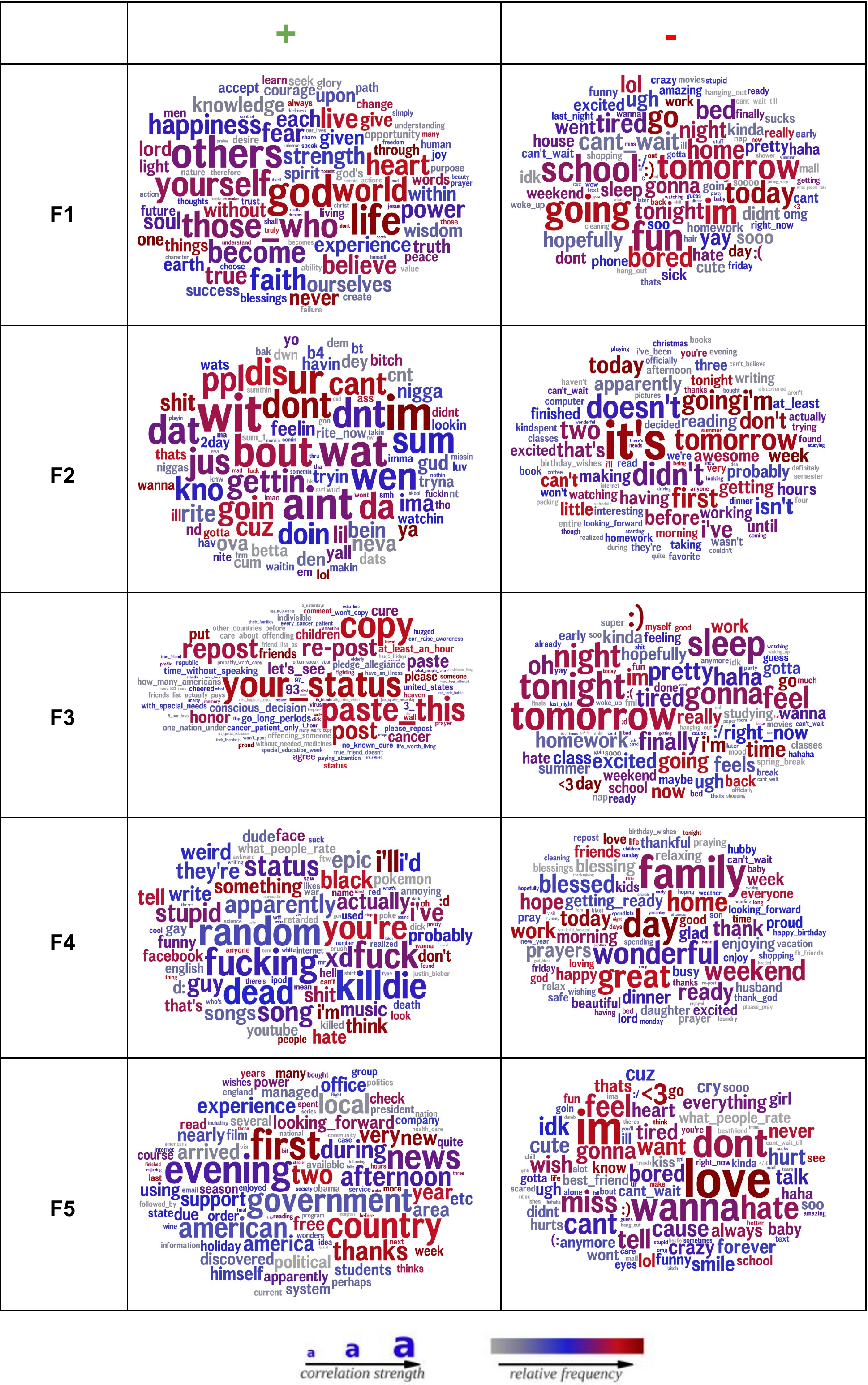}
\caption{\small Word clouds showing the most/least correlated words for each FA factor (with rotation) as obtained using Differential Language Analysis (\cite{schwartz2013personality}). %\cite Schwartz et al., 2013%
The larger the word, the more strongly it correlates with the factor. Color indicates frequency (grey = low use, blue = moderate use, red = frequent use).} 
\label{fig:wordclouds_corr}
\end{figure*}
\fi

\section*{Differential Analysis and Convergent Validity}
We next explore how our derived factors relate with one another and with the \textsc{Big5}, using Pearson correlation coefficients (r). If the factors are capturing real personality characteristics, they should be correlated to some extent with the personality factors that have dominated personality research. But if they capture unique variance, then correlations should only be low to moderate. They should also minimally correlate with one another. As illustrated in Figure \ref{fig:convergent_validity}, the factors most consistently correlate with extraversion and openness. F4 demonstrates the strongest correlation, reflecting lower levels of conscientiousness. The F4 words are similar to \cite{kern2014online}, with numerous swear words on the positive side and family, work, and relaxation words on the negative side \cite{kern2014online}. %Kern et al., 2014 personality paper

\ifdefined\submit
\subsection*{Fig 2. Correlations between the learned factors and the Big5 factors.}
\refstepcounter{section}
\label{fig:convergent_validity}
\else
\begin{figure*}[ht!]
    \centering
        \includegraphics[width=0.9\textwidth]{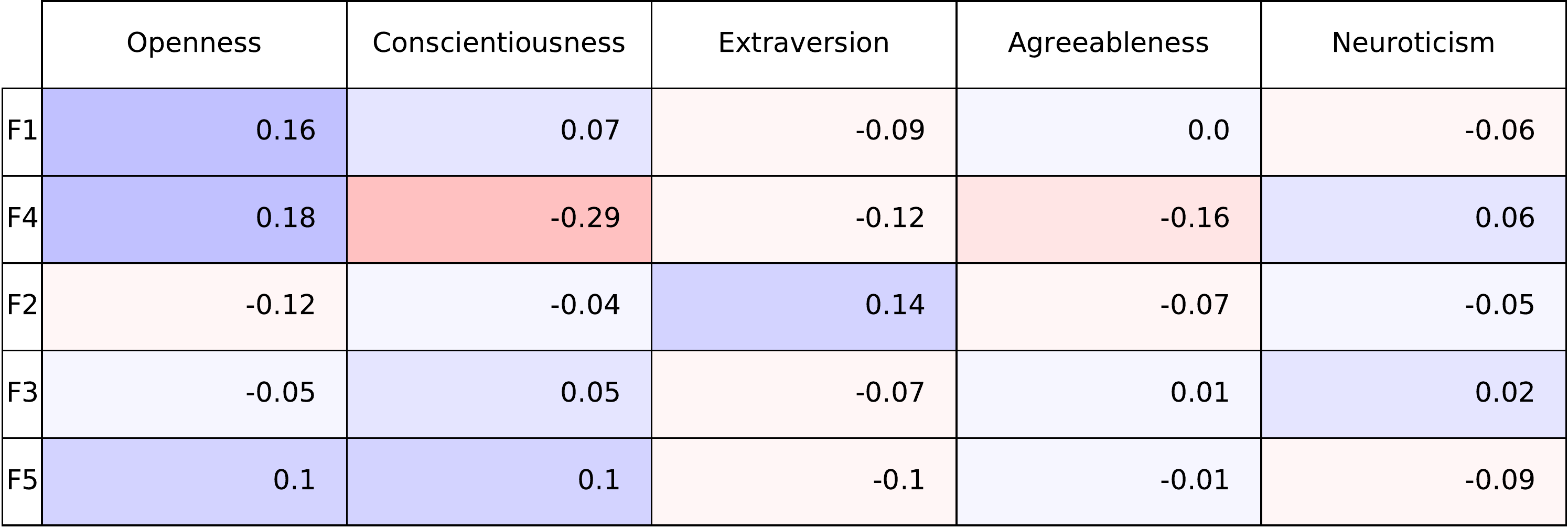}
   \caption{\small Correlations between the learned factors and the \textsc{big5} factors. The factors have been rearranged to highlight a strong diagonal for easier interpretation.}
\label{fig:convergent_validity}
\end{figure*}
\fi

We also report the correlation of each individual factor with outcomes which we show in Figure \ref{fig:individual_fa_corr}. In general, we note that each individual factor correlates slightly with outcomes but we also observe stronger correlations. Observe for example, that Factor F2 which appears to capture the behavior of teenagers correlates negatively with income. Furthermore, observe how Factor F4, which captures people who are offended (and use swear words) is negatively correlated with Satisfaction with Life (SWL).

\ifdefined\submit
\subsection*{Fig 3. Individual Factor Correlations with Outcomes.}
{\small  Note how F4 which captures the use of swear words negatively correlates with Satisfaction with Life (SWL).}
\refstepcounter{section}
\label{fig:individual_fa_corr}
\else
\begin{figure*}[htb!]
    \centering
        \includegraphics[width=0.9\textwidth]{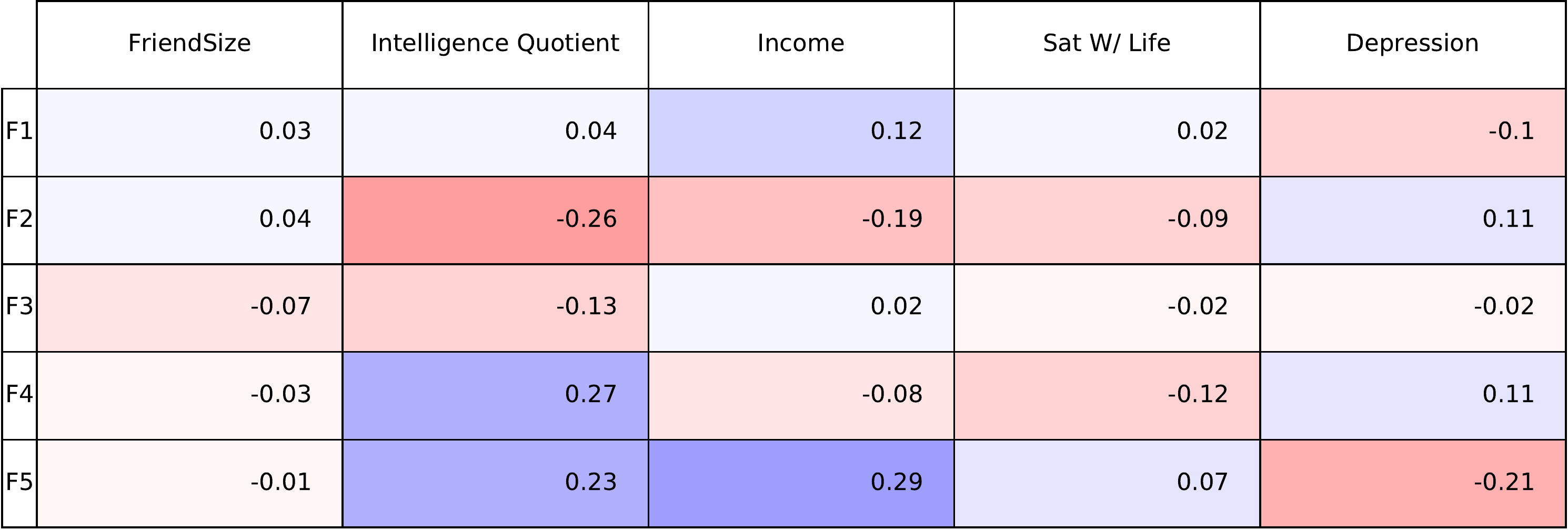}
   \caption{\small Individual Factor Correlations with Outcomes. Note how F4 which captures the use of swear words negatively correlates with Satisfaction with Life (SWL).}
\label{fig:individual_fa_corr}
\end{figure*}
\fi

We now show the $2$ \textsc{Big5 Questions} and the $5$ \textsc{Likes} that  correlate the most positively and negatively with each individual factor in Figure \ref{fig:infograph} which reveal some insights into the human behaviors captured by our factors. To give a couple of examples, Factor F1 captures traits of people who have a rich vocabulary and like to engage in deeper conversations. Note that people with these traits also like philosophy and like \textsc{Dalai Lama, The Alchemist and TED}. On the other hand, people with a low factor F1 score are not interested in theoretical discussions, have difficulty understanding abstract ideas, and like animated movies such as \textsc{Finding Dory}. Similarly note that factor F5 also captures openness and a liberal outlook, where people who have a rich vocabulary tend to vote for liberal political parties and watch/listen to \textsc{NPR}, \textsc{PBS} and \textsc{The Daily Show}. 

\ifdefined\submit
\subsection*{Fig 4.  \textsc{Questions} (left of each factor) and \textsc{Likes} (right of each factor) that correlate the highest (green) and lowest (pink) for each of our 5 behavioral-linguistic trait factors.}
\refstepcounter{section}
\label{fig:infograph}
\else
\begin{figure*}[hb!]
    \centering
        \includegraphics[angle=90, width=\textwidth,height=\textheight,keepaspectratio]
{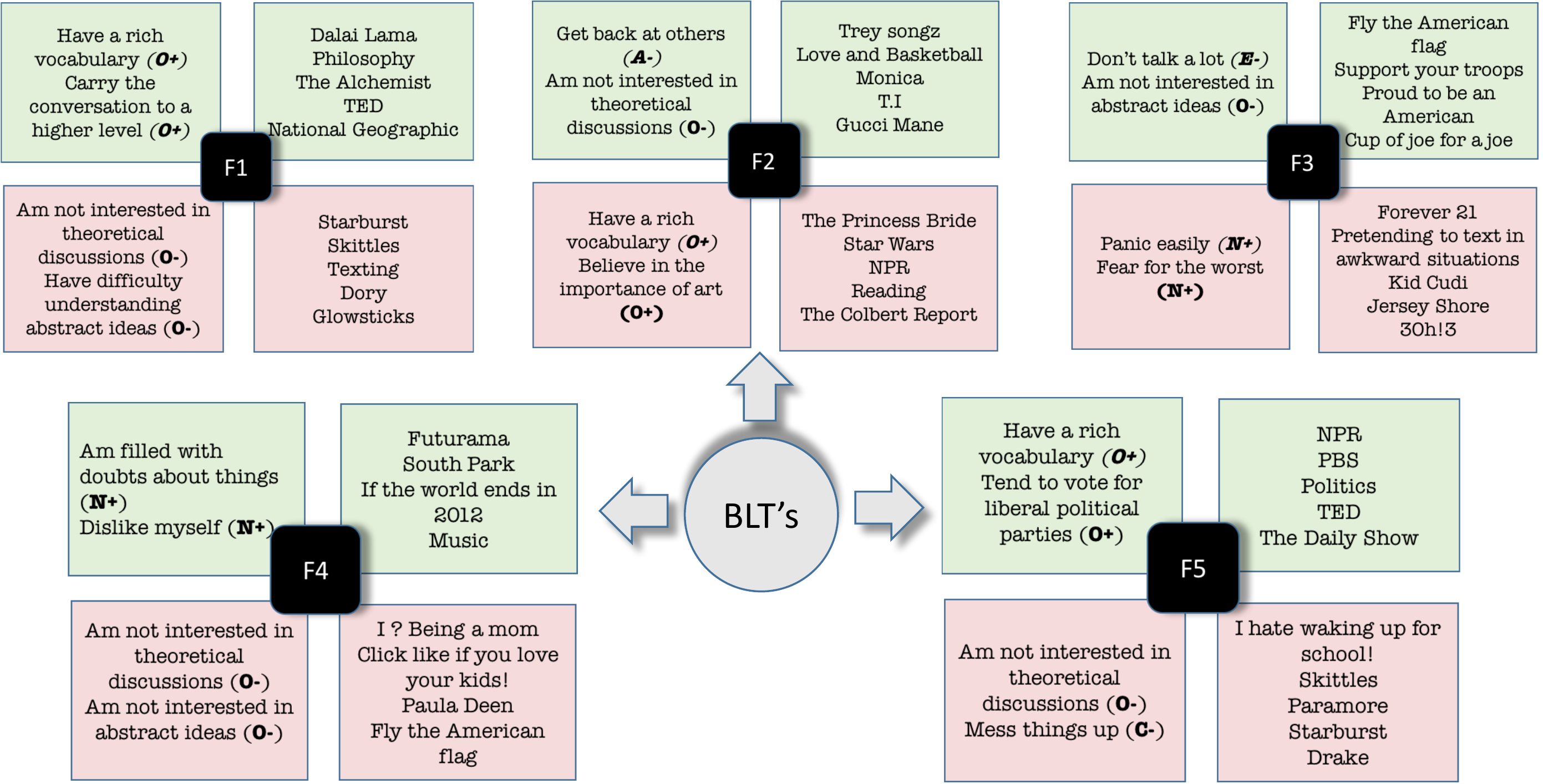}
   \caption{\small \textsc{Questions} (left of each factor) and \textsc{Likes} (right of each factor) that correlate the highest (green) and lowest (pink) for each of our 5 behavioral-linguistic trait factors.}
\label{fig:infograph}
\end{figure*}
\fi

In summary, all of these examples suggest that our learned factors capture a variety of human behavior as observed on social media, entirely derived from their language using an open vocabulary approach.

Finally, Figure \ref{fig:wordclouds_rotated_effect} illustrates the effect of rotation on the factor structure. While the un-rotated version has multiple factors that are characterized by words like ``paste this'' and ``status update'', note the stark absence of words like ``paste this'' in the rotated version in multiple factors thus yielding a more distinct factor structure.

\ifdefined\submit
\subsection*{Fig 5. Word clouds showing the effect of a rotation.}
{\small Word clouds showing the effect of a rotation. A rotation yields markedly distinct factors. Note the absence of words like ``paste this'' in the rotated version in multiple factors as opposed to the unrotated version where multiple factors are characterized by words like ``paste this'' and ``status update''. 
The larger the word, the more strongly it correlates with the factor. Color indicates frequency (grey = low use, blue = moderate use, red = frequent use) \cite{schwartz2013personality}}.
\refstepcounter{section} 
\label{fig:wordclouds_rotated_effect}
\else
\begin{figure*}[htb!]
\centering
\includegraphics[clip, width=\textwidth]{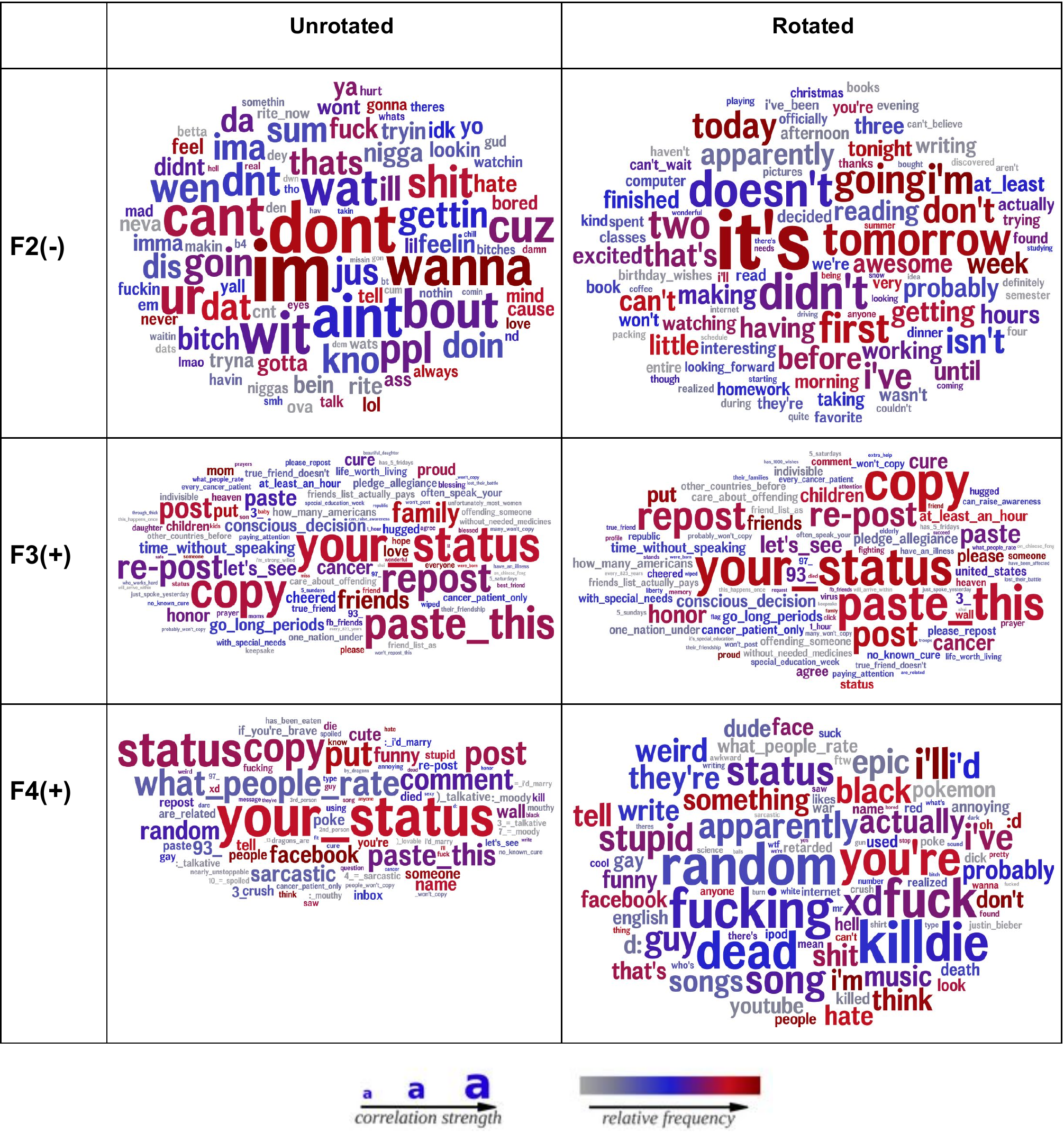}
\caption{\small Word clouds showing the effect of a rotation. A rotation yields markedly distinct factors. Note the absence of words like ``paste this'' in the rotated version in multiple factors as opposed to the unrotated version where multiple factors are characterized by words like ``paste this'' and ``status update''. 
The larger the word, the more strongly it correlates with the factor. Color indicates frequency (grey = low use, blue = moderate use, red = frequent use) \cite{schwartz2013personality}.}
\label{fig:wordclouds_rotated_effect}
\end{figure*}
\fi

\section*{Evaluation}
\label{sec:fac_eval}
In this section, we present methods to comprehensively evaluate our derived factors. Our evaluations broadly seek to quantify two aspects of the factors: \emph{generalizability} and \emph{stability}. 

\subsection*{Generalizability}
%PK – this section kept repeating the task that was done, but doesn’t say where these measures come from or number of users with this information. I put the repetitive information together and noted places where the source is needed.
Predictive validity seeks to measure the \emph{generalizability} of the learned factors by measuring their predictive performance on a number of tasks, based on other data available on the same users. We group these evaluations into two categories: \emph{questionnaires/survey} and \emph{behavioral/ economic} outcomes. For each outcome, we predict the outcome using regression, and report our performance using the Pearson r correlation coefficient (except in the case of categorical classification where we report AUC).

\paragraph*{Behavioral/Economic Outcomes}
\begin{enumerate}
\item \textsc{FriendSize}: The number of friends was pulled from users' Facebook profiles. 
\item \textsc{Income}: Estimated income was available through a 5 minute long questionnaire administered to $2,623$ Facebook users in the US which includes questions on age, gender, educational qualifications and income. Due to the skewed distribution of income, we use the logarithm of reported income. 
\item \textsc{Intelligence Quotient (IQ)}: Users completed an IQ test through the MyPersonality platform \cite{kosinski2015facebook} 
\item \textsc{Likes}: We predict a small number of broad categories that a users like, such as \textsc{Rock Music Bands}, \textsc{Gaming}, and \textsc{Hobbies}. To determine these categories, we began with a matrix $N$ of Users and  fine-grained categories of likes on Facebook. We considered only the top $10,000$ likes by popularity (i.e., frequency counts across users). We then clustered the users with Non-negative matrix factorization (NMF) to reduce the dimensionality of $N$ to $20$ clusters (which we evaluated qualitatively). As an illustration, we show below one such broad level cluster which corresponds to music bands in the metal genre of music: %pk indicate how you chose 20 clusters. Is it random? Doesn’t matter why, just note rationale
\begin{framed}
\texttt{\textbullet \; Disturbed \textbullet \; System of a Down \; \textbullet \;  Linkin Park\\
\indent
\textbullet \;  Slipknot \textbullet \; Avenged Sevenfold \textbullet \; Breaking Benjamin\\
\indent 
\textbullet \; Bullet for my Valentine \textbullet \; Metallica \textbullet \; Korn 
}
\end{framed}
\end{enumerate}

\paragraph*{Questionnaire/Survey Outcomes}
\begin{enumerate}
\item \textsc{Life Satisfaction (SWL or Sat. W/ Life)}: Satisfaction with life (SWL) was obtained through the 5-item Satisfaction with Life questionnaire \cite{diener1985satisfaction}. % cite Diener et al., 1985 Diener, E., Emmons, R. A., Larsen, R. J., & Griffin, S. (1985). The satisfaction with life scale. Journal of Personality Assessment, 49(1), 71-75. 
\item \textsc{Depression Ratings}: Depression was obtained through the 20-item Center for Epidemiological Studies-Depression (CES-D) questionnaire \cite{radloff1977ces}. %cite Radloff 1977. Radloff, L. S. (1977). The CES-D Scale: A self-report depression scale for researchi n the general population. Applied Psychological Measurement, 1, 385-401. 
\item \textsc{Additional Big 5 Questions}: Beyond the 20 personality items used to calculate the  \textsc{big5} scores, many users had between 10 and 80 additional personality items available. The evaluative task is to predict personality based on these additional items. We consider only Questions $21-100$, since the first $20$ questions were directly used to compute their \textsc{Big5} scores. This, thus examines predictive validity over and above basic  \textsc{Big5} correlations.
\end{enumerate}

\paragraph{Learning Predictive Models}
For regression tasks we use Linear Regression with $L2$ penalty (Ridge Regression), and for  classification tasks we  train a Logistic Regression classifier. In both cases, we restrict analyses to linear models to ensure that our models are interpretable and reveal the inherent predictive power of the factors. We  set  our hyper parameters using a grid search and use cross-validation to provide more stable estimates and to quantify model variance. We report our results  as the mean performance over 10 different random splits of training and test data. 

%\subsection*{Test/Retest Validity}
\subsection*{Stability}
\paragraph*{Test/Retest validity}
We evaluate the stability of the learned factors by conducting a test-retest experiment. Our experimental procedure is as follows:
\begin{enumerate}
\item Split the entire corpus of messages into two parts: (a) a training portion used for learning a model to infer factors ($75\%$ of the corpus) and (b) a test portion ($25\%$ of the corpus) which is held out from training to test the estimated model. We further divide each user's posts into several time periods (6 months apart). 
\item Learn a model to infer factors for each user using the training set. 
% is this different than the model that was learned above? Why learn another one?
% vivek We learned the factors on the whole data set for predictive validity. This helps us to evaluate predictive validity on multiple splits etc and use more data for fitting a model. Finally in order to demonstrate test-retest validity we split the entire time period into 5 time periods where we use only 1 time period of inferring factors and show stability across the remaining.
\item Infer factors for users in each time period of the test set.
\end{enumerate}
We report the correlations across available time periods. Park et al. \cite{park2015automatic} found that language-based assessments of the \textsc{Big5} factors demonstrated strong correlations over time, ranging from r = .62 for neuroticism to r = .74 for openness across consecutive 6 month intervals, with lower correlations across farther time periods \cite{park2015automatic}. If the BLT factors demonstrate endurance, then we'd expect to see strong correlations on the test set (r = .60 and above) across a 6 month interval, with some declines across subsequent temporal periods. 
% pk - are the correlations based on the test set, the training, or a combination? I'm not clear on this
%We discuss our results on  test-retest validity in Section \ref{sec:results_testretest}.
%vivek: Clarified
\paragraph*{Dropout Reliability} Finally, we  evaluate the external validity of the learned factors across different  samples  of users.  The learned factors should not be overly dependent on a specific set of users from the training data.  We quantify the sensitivity of the learned factors to the presence (or absence) of users as  follows:
\begin{enumerate}
\item We randomly drop $20\%$ users from the training data before we learn the factors.
\item We repeat step 1 one hundred times.  
\item We  infer factor scores on the fixed held out test set for each of the 100 learned models.
\item We consider the factor scores inferred from each pair of models $(i,j)$:
     \begin{itemize}
      \item We use the Hungarian Algorithm \cite{kuhn2010hungarian} to infer the best alignment of factor scores, as measured by correlations.  % what is this Algorithm? CItation? %vivek: Done
     \item We compute and report the mean correlation between the aligned factor scores.
     \end{itemize}
\item We do this for each model pair and compute the mean of the scores observed.
\end{enumerate}
%We report the results of Dropout reliability in Section \ref{sec:results_testretest}.

% Results and Discussion can be combined.
%\section*{Results}
\section*{Results and Discussion}
\label{sec:results}
%\subsection{Correlation Structure}
%\label{sec:results_convergent}
\begin{comment}
\begin{figure*}[t!]
    \centering
    \begin{subfigure}[t]{0.45\textwidth}
        \centering      
        \includegraphics[width=\textwidth]{lower_centered/lsi_not_residualized_rotated_corr_lower.pdf}
  \caption{SVD}
  \end{subfigure}
  \begin{subfigure}[t]{0.45\textwidth}
        \centering      
        \includegraphics[width=\textwidth]{lower_centered/fa_not_residualized_rotated_lower.pdf}
  \caption{FA}
  \end{subfigure}
    \caption{Correlations of learned factors (\textbf{only lower triangle shown}) with \textsc{big5} after applying a \textsc{Varimax} rotation which seeks to maximize the variance of loadings. One implication of this is that rotated loadings are sparse and hence more interpretable. For example, note how factor F2 in SVD correlated well with \textsc{openness} and factor F5 correlates with \textsc{conscientiousness}.}
\label{fig:rotate_corr}
\end{figure*}
\end{comment}

\subsection*{Generalizability}
\label{sec:results_predictive}
Table \ref{tab:predno_res} shows the predictive validity of the \textsc{Big5} factors, or \textsc{BLTs} (FA5), and models that include age and gender, reported as the mean Pearson r correlation coefficient over 10 random train-test splits for each outcome.
% I don't understand what you mean by residualizing demographics - in the table you present it with and without demographics. How is that different from residualizing the demographics? And what is meant by demographics? Is that age and gender? 
% vivek yes demographics here means age and gender. 
% discuss the questionnaires first, to mirror the way they were presented in the method section. And not clear why they are called different things here than in the method. I've changed to be consistent across sections. Also, I would make this a single table of outcomes, not two tables.
For questionnaire-based outcomes, the \textsc{Big5} factors better predict SWL and depression than our factors. This is driven by strong correlations of SWL and depression with extraversion and neuroticism. The \textsc{Big5} also outpredicts our factors for the size of the friend network, which again is driven by strong correlations with extraversion. % I look at this and immediately wonder what is driving the results - I made up these sentences with the results that I would expect. Would be good to have this broken down for each big 5 and FA5 factor in the Supporting Information. In personality research, we never look at these as just a single model - we look at the patterns for the 5 factors, and I think we should be doing that here as well, at least in the supplement.  I also find it really interesting that predictive validity is euqivalent for the big5 questions. 
Notably, correlations are equivalent for the \textsc{Big5} and \textsc{FA5} in predicting the additional personality questions (Big5Questions). In contrast, as highlighted in the table, our factors better predict income, IQ, and the user likes.  
%First, we discuss results for the social/demographic outcomes. Table \ref{tab:social_predno_res} shows our performance on these outcomes using FA factors.  
%Observe that on these outcomes, our language based factors outperform questionaire based factors like \textsc{big5} (highlighted with \colorbox{high}{COLOR} in Table \ref{tab:social_predno_res}).
For example, on the task of predicting \textsc{Likes}, FA with 5 factors outperforms the baseline by $7$ percentage points (60.11 - 52.6).  
%whereas FA with $10$ factors reveals an increase of $11\%$. 
It is worth emphasizing here that the task of predicting \textsc{Likes} is a challenging task - it  essentially involves $20$ different classification tasks. Consequently, an improvement of $7$ percentage points on this task is  promising. Also note that adding age and gender as covariates improves predictive performance (compare \textsc{FA5+Demog} with \textsc{FA5}). 
%Table \ref{tab:social_pred_res} shows the performance on these set of outcomes using factors with demographics (age and gender) residualized. While we observe that residualizing out demographics generally reveals a drop in performance of $5\%$ over the \textsc{Likes} outcome (see rows \textsc{FA5}, \textsc{FA10} and \textsc{FA25} for \textsc{Likes} column in Tables \ref{tab:social_predno_res} and \ref{tab:social_pred_res}), nonetheless FA based factors consistently outperform the \textsc{big5} based factors on predicting behavioral/economic outcomes.

%Now we turn our attention to results on questionnaire based factors as shown in Tables \ref{tab:questionnaires_predno_res}. First note that FA based factors perform competitively with \textsc{big5} on the task of \textsc{big5questions} where \textsc{big5} has an inherent advantage since these questions are correlated with \textsc{big5} scores by design. Note also that on the tasks of \textsc{SWL} and \textsc{Depression}, language based factors (FA) do not perform as well as \textsc{big5}. We hypothesize two reasons for this under-performance: (a) Language based factors do not capture very strong psychological variables like depression etc. very well and (b) Questionnaire based methods are subject to shared method variance which is manifested by a higher correlation of these variables with respect to the \textsc{big5}. 

% put table here (single table, not two parts)

%%% PK make this a single table. I adjusted the caption for the whole table, but can't figure out how to create the table without getting errors
\begin{table*}[t!]
	\small
	\centering
\begin{subtable}{\linewidth}
	\small
	\centering
	\begin{tabular}{l|c|c|c|c}
		\textbf{Method} & \textsc{FriendSize} & \textsc{Income} & \textsc{IQ} & \textsc{Likes} \\
        \hline
        \textsc{Demog} & 0.052 & 0.283 & 0.162 & 55.5 \\
        \textsc{Big5} &  0.183 & 0.037 & 0.179 & 52.6 \\
        \textsc{Big5}+\textsc{Demog} & 0.192 &  0.278 & 0.269 & 56.9  \\
        \hline
        FA5 & 0.125 &  \cellcolor{high}0.362 & \cellcolor{high}0.361 & \cellcolor{high}60.11\\
        %FA10 & 0.178 & \cellcolor{high}0.378 & \cellcolor{high}0.395 & \cellcolor{high}63.39 \\
        %FA25 & \cellcolor{high}0.312 & \cellcolor{high}0.363 & \cellcolor{high}0.390 & \cellcolor{high}64.76 \\
        \hline
        FA5 + \textsc{Demog} & 0.148 &  \cellcolor{high}0.375 & \cellcolor{high}0.423 & \cellcolor{high}61.86\\
        %FA10 + \textsc{Demog} & \cellcolor{high}0.191 & \cellcolor{high}0.411 & \cellcolor{high}0.396 & \cellcolor{high}64.32 \\
        %FA25 + \textsc{Demog} & \cellcolor{high}0.323 & \cellcolor{high}0.391 & \cellcolor{high}0.416 & \cellcolor{high}65.54 \\
 	\end{tabular}
    \caption{\small \textbf{Behavioral/Economic Outcomes}: We show mean Pearson's R over 10 random train-test splits for \textsc{FriendSize}, \textsc{Income} and \textsc{IQ} while for \textsc{Likes} we show the mean area under the curve (AUC) over all $20$ categories. Language based factors (FA) perform competitively and even outperform questionnaire based factors as highlighted in color.}
\label{tab:social_predno_res}
\end{subtable}
\begin{subtable}{\linewidth}
	\small
	\centering
	\begin{tabular}{l|c|c|c}
		\textbf{Method} & \textsc{Big5Questions} & \textsc{Sat. W/ Life} & \textsc{Depression} \\
        \hline
        \textsc{Demog} & 0.072 & 0.053 & 0.103 \\
        \textsc{Big5} & 0.178 & 0.486& 0.407  \\
        \textsc{Big5}+\textsc{Demog} & 0.191 &  0.524 & 0.424  \\
        \hline
        FA5 & 0.178 &  0.165 & 0.293 \\
        %FA10 & \cellcolor{high}0.191 & 0.229 & 0.256 \\
        %FA25 & \cellcolor{high}0.213 & 0.231 & 0.298 \\
        \hline
        FA5 + \textsc{Demog} & 0.186 &  0.207 & 0.227 \\
        %FA10 + \textsc{Demog} & \cellcolor{high}0.199 & 0.241 & 0.260 \\
        %FA25 + \textsc{Demog} & \cellcolor{high}0.218 & 0.249  & 0.298 \\
 	\end{tabular}
	\label{tab:questionnaires_predno_res}
    \caption{\small \textbf{Questionnaire based outcomes} We show mean Pearson R over 10 random train-test splits. Language based factors (FA) do not outperform questionnaire based factors.}
\end{subtable}
\caption{\small \textbf{Questionnaire and Behavioral/Economic Outcomes}: Predictive performance over 10 random train-test splits for the questionnaire, behavioral, and economic outcomes, for the Big5 factors, our learned language based factors (FA5), and demographics (age and gender; DEMOG).}
\label{tab:predno_res}
%vivek: @Peggy, We decided to clearly demarcate both categories of predictive tasks. So keeping both tables.
\end{table*}

Tables  \ref{tab:question_perf} and \ref{tab:likes_perf} show the best (top) and worst (bottom)  \textsc{Big5} items and \textsc{LIKES}, ranked according to the predictive performance by our  \textsc{BLTs}. For \textsc{Big5} questions, \textsc{BLTs} best predicted openness items, and less predictive of extraversion and conscientiousness items. The strongest items focus on language (e.g.,  \emph{``have a rich vocabulary''} (see Table \ref{tab:question_perf_best}), suggesting that the \textsc{BLTs} are adequately capturing everyday behavior. For \textsc{LIKES}, \textsc{BLTs} predict general categories (e.g., music genres, children focused), and is least predictive of generic \textsc{LIKE}s which almost all users might like (for example: \textsc{Youtube, Facebook}).    

We conclude by emphasizing that the traits learned using FA are \emph{not apriori tuned} to any particular predictive task, and yet perform competitively with traits derived from questionnaires in predicting a variety of outcomes and even outperform questionnaire based traits on behavioral outcomes like \textsc{Income} and \textsc{IQ}, thus underscoring the generalizability of these traits.

\begin{table*}[t!]
	\small
	\centering
\begin{subtable}{\linewidth}
	\small
	\centering
	\begin{tabular}{l|c|c}
		\textbf{QNO} & \textsc{Question} & \textsc{R}  \\
        \hline
        \textsc{54} & \emph{Am not interested in theoretical discussions} (O-) & 0.230  \\
        \textsc{71} & \emph{Have a rich vocabulary} (O+) & 0.224 \\
        \textsc{64} & \emph{Have difficulty understanding abstract ideas} (O-) & 0.222  \\
         \textsc{51} & \emph{Tend to vote for liberal political candidates} (O+) & 0.220 \\

       \textsc{90} & \emph{Am filled with doubts} (N+) & 0.215 \\

        \hline
 	\end{tabular}
 	\caption{5 questions that the \textsc{BLTs} best predict.}
	\label{tab:question_perf_best}
\end{subtable}
\begin{subtable}{\linewidth}
	\small
	\centering
	\begin{tabular}{l|c|c}
		\textbf{QNO} & \textsc{Question} & \textsc{R}  \\
        \hline
        \textsc{28} & \emph{Waste my time} (C-) & 0.094  \\
        \textsc{63} & \emph{Am the life of a party} (E+) & 0.131  \\
        \textsc{43} & \emph{Talk to a lot of different people at parties} (E+) & 0.133  \\
        \textsc{29} & \emph{Dont talk a lot} (E-) & 0.135  \\
        \textsc{88} & \emph{Find it difficult to get down to work} (C-) & 0.139  \\
        %\textsc{75} & \emph{Get chores done right away} (C+) & 0.149  \\
        \hline
 	\end{tabular}
 	\caption{5 questions that the \textsc{BLTs} least predict.}
	\label{tab:question_perf_worst}
\end{subtable}
\caption{The Big5 questions on which the \textsc{BLT} factors do the best (top) and worst (bottom) at predicting. }
\label{tab:question_perf}
\end{table*}

\begin{table*}[t!]
	\small
	\centering
\begin{subtable}{\linewidth}
	\small
	\centering
	\begin{tabular}{l|p{8cm}|c}
		\textbf{LIKENO} & \textsc{Like} & \textsc{AUC}  \\
        \hline
        \textsc{8} & \emph{Lady Antebellum, Tim McGraw, NCIS, Kenny Chesney, Country music, Jason Aldean, Walmart, Carrie Underwood, George Strait, Family Feud}
 & 71.04  \\
 \textsc{12} & \emph{Glowsticks, Finding Nemo, Being Hyper!, DORY} & 68.50  \\
 \textsc{11} & \emph{Lil Wayne, Drake, Eminem,T.I., Nicki Minaj, Jersey Shore,Trey}  & 68.50  \\
 \textsc{10} & \emph{I redo high fives if they weren't good enough the first time
,Why do we have to be quiet during a fire drill? Will the fire hear us?} & 68.05  \\
 \textsc{14} & \emph{The Beatles, Pink Floyd, The Doors, Radiohead, Queen, Nirvana}
& 65.85  \\
        \hline
 	\end{tabular}
    \caption{5 Likes that the \textsc{BLTs} best predict.}
	\label{tab:likes_perf_best}
\end{subtable}

\begin{subtable}{\linewidth}
	\small
	\centering
	\begin{tabular}{l|p{8cm}|c}
		\textbf{LIKENO} & \textsc{Like} & \textsc{AUC}  \\
        \hline
 \textsc{3} & \emph{I hate when im yelling at someone and i mess up what im saying, I would take a bullet for u.. Not the head but like in the leg or something}
& 50.73  \\
        \textsc{0} & \emph{YouTube, Facebook, Oreo, Skittles, Coca-Cola, Adam Sandler, Starburst, Starbucks, Music, Toy Story} & 51.46  \\
\textsc{16} & \emph{After an arguement I think about clever things I should have said, Your in a good mood, one little thing happens and BAM....  Bad mood.} & 54.23  \\
        \textsc{4} & \emph{I love days in class when all we do is chill and talk the whole time
Get real. No one's going to form a single line if the building's on FIRE.} & 54.49  \\
        \textsc{1} & \emph{When I was little I liked building forts out of pillows and blankets, I like when my scissors glide through the paper so I don't have to cut.}
& 54.78  \\
        \hline
 	\end{tabular}
 	\caption{5 Likes that the \textsc{BLTs} least predict.}
	\label{tab:likes_perf_worst}
\end{subtable}
\caption{The Categories of Likes that the \textsc{BLT} factors do the best (top) and worst (bottom) at predicting. We show the top \textsc{LIKES} from the clusters for interpretation. AUC = area under the curve}
\label{tab:likes_perf}
\end{table*}

\subsection*{Stability}
\label{sec:results_testretest}
Figure \ref{fig:test_retest} shows the factor correlations at subsequent time periods with the factor scores at the initial point ($t = 0$) over a common set of users. Four of the five factors demonstrate strong correlations, with some decline over subsequent periods, but generally are fairly stable. The exception is F3. The words in this factor (see Figure \ref{fig:wordclouds_corr}) reflect Facebook behavior (e.g., paste, your status, repost) versus offline behavior (e.g., tomorrow, tonight, sleep, excited). 
In this case, we believe this factor is capturing less of a trait and more of a temporary ``new Facebook user'' state. 
%The lack of stability may be capturing the dynamic aspects of personality, which are known to vary across situations \cite{mischel2004toward}. % cite Mischel 2004
% Mischel, W. (2004). Toward an integrative science of the person. Annual Review of Psychology, 55, 1-22.
\ifdefined\submit
\subsection*{Fig 6. Test re-test validity of our learned factors.}
\refstepcounter{section}
\label{fig:test_retest}
\else
\begin{figure*}[htb!]
    \centering
        \includegraphics[width=\textwidth]{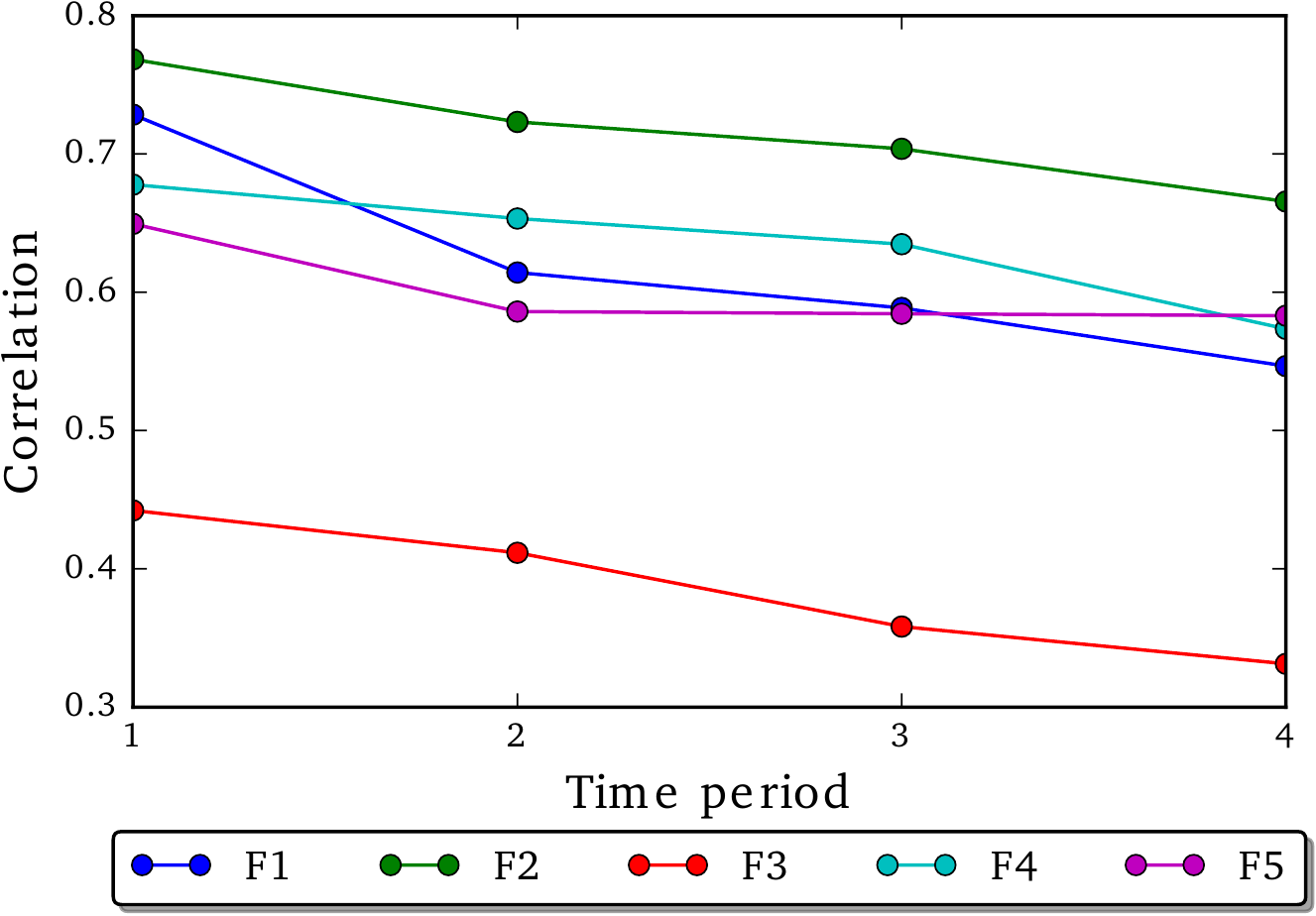}
    \caption{\small Test-retest validity: Correlations of factors observed over different time periods with the factors at time $t=0$ measured over the same set of users in a test-retest setting.}
 \label{fig:test_retest}
\end{figure*}
\fi

%Personality is both stable and variable, and our BLTs appear to capture both of these elements as they are reflected in everyday language.

Finally, for the dropout analysis we computed the mean correlation among factors obtained over multiple runs where a random sample of $20\%$ were dropped before learning the factors. We used the Hungarian algorithm to infer the  mapping between factors across multiple runs. There was a very strong average correlation  ($0.94$) among corresponding factors across multiple runs. 
% PK I may be misinterpretting what is noted in the method section, but step 5 noted reporting a distribution of scores. Seems like that should be reported? Or the exact average correlation should be reported here.
% vivek: Fixed reported the exact average.

Both the test-retest analysis and dropout analysis supports the endurance of the factors across  time and sub-populations.

\section*{Limitations}
In this first exploration of deriving traits from human behavior manifest on social media, we left many questions unanswered. 
Our latent traits capture human behavior as revealed in the context of Facebook over $49,139$ US users which may not be generalizable across the globe. 
We leave it to future work to explore whether similar traits are observed in other contexts. 
Furthermore, while traditional methods suffer from response biases, our methods have others -- for example, although Facebook is used by a majority of Americans there is still no doubt some selection biases in who is on it and people may be projecting how they want to be seen. % as than rather than who they actually are.
We see such an approach as a supplement to traditional techniques. 
Finally, we did not directly address interpretability of our learned factors and did not explicitly impose stability constraints (and revealed we had one more dynamic state-like factor). 
On the other hand, our goal has been to determine the feasibility of deriving traits from social media behavior, rather than to propose a specific set of traits, akin to the big five, to be used across all contexts. 
We leave it to future work to rigorously analyze and address these aspects.

\section*{Conclusion}
\label{sec:conclusion}
We proposed a method based on \textbf{factor analysis} to infer latent personality traits from the everyday language of users on social media. 
A person is "an organized, dynamic, agentic system functioning in the social world", with characteristics, behaviors, feelings, and thoughts that are both consistent and variable across time and situations \cite{mischel2004toward}. % Mischel 2004, p. 2
Individual differences have often been based on questionnaire items, which may not capture the richness of everyday human behavior. 

%Rather than relying on manually curated items, 
Our method infers latent factors from linguistic behavior in social media -- a medium that allows access to large sample sizes; unprompted access to user's thoughts, emotions, and language; and data-driven approaches.  
We derived five factors from the language, and demonstrated that these factors are \emph{generalizable} with good predictive power, and \emph{stable} across time and sub-populations. 
We see this as a stepping stone to deriving a supplement to the popular Big Five personality traits based on large-scale behavioral data rather than questionnaire self-reports. 
We have demonstrated it is feasible to produce factors from social media language that have predictive and face validity, stability and sometimes generalize better than the Big Five. 

%Other factors most likely exist, and subsequent work should explore factors that appear across different mediums, populations, and temporal periods. 
%We believe that this work will set future directions for large scale analysis of social media text to test psychological theories and hypotheses, enabling psychologists to gain insights into people by observing their everyday behavior at scale.
\beginsupplement
\section*{Supporting Information}
% PK I really think we need a table that shows the predictive results for each of the Big5 factors and each of the FA5 factors. It seems odd to me to just group them all together into a single thing, and seems like it covers over some of the explanation for the results that are observed. 
In this section, we  provide supporting results based on our analyses of BLTs. Table \ref{tab:predno_res_5_30} shows the performance of extracting 10 and 30 factor BLTs (FA10 and FA30) on both categories of predictive tasks outlined (to align with the ten aspects and 30 facets that underlie the Big5 factors). For comparison, with the questionnaire items, we calculate the 10 aspect scores and 30 facet based scores, using the relevant IPIP items. BLTs consistently outperform questionnaire based models on behavioral/demographic outcomes, and underperform on questionnaire based outcomes. Similar results are also obtained with demographics residualized (see Tables \ref{tab:pred_res} and \ref{tab:pred_res_5_30}).
% again, make the questionnaire & behavioral/ economic outcomes a single table, then a single table for residulaized results. Adjust the captions accordingly (I did not do that here)

\begin{table*}[htb!]
	\small
	\centering
\begin{subtable}{\linewidth}
	\small
	\centering
	\begin{tabular}{l|c|c|c|c}
		\textbf{Method} & \textsc{FriendSize} & \textsc{Income} & \textsc{IQ} & \textsc{Likes} \\
        \hline
        \textsc{Demog} & 0.052 & 0.283 & 0.162 & 55.50 \\
        \textsc{Big5-10} &  0.202 & 0.147 & 0.252 & 53.50 \\
        \textsc{Big5-10} +\textsc{Demog} & 0.200 &  0.344 & 0.220 & 57.30  \\
        FA10 & 0.178 & \cellcolor{high}0.378 & \cellcolor{high}0.395 & \cellcolor{high}63.39 \\
        FA10 + \textsc{Demog} & \cellcolor{high}0.191 & \cellcolor{high}0.411 & \cellcolor{high}0.396 & \cellcolor{high}64.32 \\
        \hline
        \textsc{Big5-30} &  0.244 & -- & 0.285 & 56.28 \\
        \textsc{Big5-30} +\textsc{Demog} & 0.233 &  -- & 0.330 & 59.32  \\
        %\hline
        %FA5 & 0.125 &  \cellcolor{high}0.362 & \cellcolor{high}0.361 & \cellcol
        FA30 & \cellcolor{high}0.316 & \cellcolor{high}0.379 & \cellcolor{high}0.420 & \cellcolor{high}64.98 \\
        %\hline
        %FA5 + \textsc{Demog} & 0.148 &  \cellcolor{high}0.375 & \cellcolor{high}0.423 & \cellcolor{high}61.86\\
        FA30 + \textsc{Demog} & \cellcolor{high}0.329 & \cellcolor{high}0.398 & \cellcolor{high}0.459 & \cellcolor{high}65.76 \\
 	\end{tabular}
  	\caption{\small \textbf{Behavioral/Economic Outcomes}: We show mean Pearson's R over 10 random train-test splits for \textsc{FriendSize}, \textsc{Income} and \textsc{IQ} while for \textsc{Likes} we show the mean area under the curve (AUC) over all $20$ categories. Language based factors (FA) perform competitively and even outperform questionnaire based factors as highlighted in color.}
\label{tab:social_predno_res_5_30}
    
\end{subtable}
\begin{subtable}{\linewidth}
	\small
	\centering
	\begin{tabular}{l|c|c|c}
		\textbf{Method} & \textsc{Big5Questions} & \textsc{Sat. W/ Life} & \textsc{Depression} \\
        \hline
        \textsc{Demog} & 0.072 & 0.053 & 0.103 \\
        \textsc{Big5-10} & 0.627 & 0.470 & 0.299  \\
        \textsc{Big5-10}+\textsc{Demog} & 0.629 &  0.479 & 0.313 \\
         FA10 & 0.191 & 0.229 & 0.179 \\
         FA10 + \textsc{Demog} & 0.199 & 0.241 & 0.222 \\
         \hline
        \textsc{Big5-30} & 0.766 & 0.583 & 0.464  \\
        \textsc{Big5-30}+\textsc{Demog} & 0.767 &  0.615 & 0.405  \\
        %FA5 & 0.178 &  0.165 & 0.236 \\
        FA30 & 0.215 & 0.229 & 0.242 \\
        %\hline
        %FA5 + \textsc{Demog} & 0.186 &  0.207 & 0.242 \\
        FA30 + \textsc{Demog} & 0.220 & 0.297  & 0.207 \\
 	\end{tabular}
 	\caption{\small \textbf{Questionnaire based outcomes} We show mean Pearson R over 10 random train-test splits. Language based factors (FA) do not outperform questionnaire based factors.}
	\label{tab:questionnaires_predno_res_5_30}
\end{subtable}
\caption{\small Predictive performance on Social media tasks and Questionnaire based tasks for factors without residualization of age and gender. \textsc{Demog} indicates that age and gender were also added as co-variates to learn predictive models.}
\label{tab:predno_res_5_30}
\end{table*}

\begin{table*}[htb!]
	\small
	\centering
\begin{subtable}{\linewidth}
	\small
	\centering
	\begin{tabular}{l|c|c|c|c}
		\textbf{Method} & \textsc{FriendSize} & \textsc{Income} & \textsc{IQ} & \textsc{Likes} \\
        \hline
        \textsc{Demog} & 0.052 & 0.283 & 0.162 & 55.50 \\
        \textsc{Big5} & 0.183 & 0.037 & 0.179 & 52.60 \\
        \textsc{Big5}+\textsc{Demog} & 0.192 &  0.278 & 0.269& 56.90  \\
        \hline
        FA5 & 0.140 &  \cellcolor{high}0.285 & \cellcolor{high}0.353& \cellcolor{high}56.33\\
        %FA10 & 0.176 & \cellcolor{high}0.309 & \cellcolor{high}0.291 & \cellcolor{high}57.85 \\
        %FA25 & \cellcolor{high}0.303 & \cellcolor{high}0.324 & \cellcolor{high}0.366 & \cellcolor{high}59.43 \\
        \hline
        FA5 + \textsc{Demog} & 0.160 &  \cellcolor{high}0.361 & \cellcolor{high}0.370 & \cellcolor{high}60.86 \\
        %FA10 + \textsc{Demog} & \cellcolor{high}0.191 & \cellcolor{high}0.360 & \cellcolor{high}0.421 & \cellcolor{high}62.10 \\
        %FA25 + \textsc{Demog} & \cellcolor{high}0.313 & \cellcolor{high}0.399 & \cellcolor{high}0.415 & \cellcolor{high}63.48 \\
 	\end{tabular}
 	\caption{\small \textbf{Behavioral/Economic outcomes}: We show mean Pearson's R over 10 random train-test splits for \textsc{FriendSize}, \textsc{Income} and \textsc{IQ} while for \textsc{Likes} we show the mean area under the curve (AUC) over all $20$ categories. Language based factors (FA) perform competitively and even outperform questionnaire based factors as highlighted in color.}
	\label{tab:social_pred_res}
    
\end{subtable}
\begin{subtable}{\linewidth}
	\small
	\centering
	\begin{tabular}{l|c|c|c}
		\textbf{Method} & \textsc{Big5Questions} & \textsc{Sat. W/ Life} & \textsc{Depression} \\
        \hline
        \textsc{Demog} & 0.072 & 0.053 & 0.103\\
        \textsc{Big5} & 0.178 & 0.486 & 0.407  \\
        \textsc{Big5}+\textsc{Demog} & 0.191 &  0.524 & 0.424  \\
        \hline
        FA5 & 0.167 &  0.232 & 0.187 \\
        %FA10 & \cellcolor{high}0.182 & 0.204 & 0.247\\
        %FA25 & \cellcolor{high}0.197 & 0.272 & 0.280 \\
        \hline
        FA5 + \textsc{Demog} & 0.185 &  0.211 & 0.289 \\
        %FA10 + \textsc{Demog} & \cellcolor{high}0.197 & 0.200 & 0.256 \\
        %FA25 + \textsc{Demog} & \cellcolor{high}0.212 & 0.277  & 0.297 \\
 	\end{tabular}
 	\caption{\small \textbf{Questionnaire based outcomes} We show mean Pearsons R over 10 random train-test splits. Language based factors (FA) do not outperform questionnaire based factors.}
	\label{tab:questionnaires_pred_res}
\end{subtable}
\caption{\small Predictive performance on Social media tasks and Questionnaire based tasks for factors with residualization of age and gender.
\textsc{Demog} indicates that age and gender were also added as co-variates to learn predictive models.}
\label{tab:pred_res}
\end{table*}

\begin{table*}[htb!]
	\small
	\centering
\begin{subtable}{\linewidth}
	\small
	\centering
	\begin{tabular}{l|c|c|c|c}
		\textbf{Method} & \textsc{FriendSize} & \textsc{Income} & \textsc{IQ} & \textsc{Likes} \\
        \hline
        \textsc{Demog} & 0.052 & 0.283 & 0.162 & 55.50 \\
        
        \textsc{Big5-10} &  0.202 & 0.147 & 0.252 & 53.50 \\
        \textsc{Big5-10} +\textsc{Demog} & 0.200 &  0.344 & 0.220 & 57.30  \\
         FA10 & 0.176 & \cellcolor{high}0.309 & \cellcolor{high}0.291 & \cellcolor{high}57.85 \\
         FA10 + \textsc{Demog} & 0.191 & \cellcolor{high}0.360 & \cellcolor{high}0.421 & \cellcolor{high}62.10 \\
         \hline
        \textsc{Big5-30} &  0.244 & -- & 0.285 & 56.28 \\
        \textsc{Big5-30} +\textsc{Demog} & 0.233 &  -- & 0.330 & 59.32  \\
        %\hline
        %FA5 & 0.140 &  \cellcolor{high}0.285 & \cellcolor{high}0.353& \cellcolor{high}56.33\\
        FA30 & \cellcolor{high}0.305 & \cellcolor{high}0.334 & \cellcolor{high}0.351 & \cellcolor{high}59.64 \\
        %\hline
        %FA5 + \textsc{Demog} & 0.160 &  \cellcolor{high}0.361 & \cellcolor{high}0.370 & \cellcolor{high}60.86 \\

        FA30 + \textsc{Demog} & \cellcolor{high}0.310 & \cellcolor{high}0.379 & \cellcolor{high}0.398 & \cellcolor{high}63.64 \\
 	\end{tabular}
 	\caption{\small \textbf{Behavioral/Economic outcomes}: We show mean Pearson's R over 10 random train-test splits for \textsc{FriendSize}, \textsc{Income} and \textsc{IQ} while for \textsc{Likes} we show the mean area under the curve (AUC) over all $20$ categories. Language based factors (FA) perform competitively and even outperform questionnaire based factors (\textsc{big5}) as highlighted in color.}
	\label{tab:social_pred_res_5_30}
    
\end{subtable}
\begin{subtable}{\linewidth}
	\small
	\centering
	\begin{tabular}{l|c|c|c}
		\textbf{Method} & \textsc{Big5Questions} & \textsc{Sat. W/ Life} & \textsc{Depression} \\
        \hline
        \textsc{Demog} & 0.072 & 0.053 & 0.103\\        
        \textsc{Big5-10} & 0.627 & 0.470 & 0.299  \\
        \textsc{Big5-10}+\textsc{Demog} & 0.629 &  0.479 & 0.313  \\
         FA10 & 0.182 & 0.204 & 0.132\\
         FA10 + \textsc{Demog} & 0.197 & 0.200 & 0.198 \\
         \hline
        \textsc{Big5-30} & 0.766 & 0.583 & 0.464  \\
        \textsc{Big5-30}+\textsc{Demog} & 0.767 &  0.615 & 0.405  \\
        FA30 & 0.197 & 0.241 & 0.127 \\      
        FA30 + \textsc{Demog} & 0.211 & 0.251  & 0.170 \\
 	\end{tabular}
 	\caption{\small \textbf{Questionnaire based outcomes} We show mean Pearsons R over 10 random train-test splits. Language based factors (FA) perform  do not outperform questionnaire based factors.}
	\label{tab:questionnaires_pred_res_5_30}
\end{subtable}
\caption{\small Predictive performance on Social media tasks and Questionnaire based tasks for factors with residualization of age and gender.
\textsc{Demog} indicates that age and gender were also added as co-variates to learn predictive models.}
\label{tab:pred_res_5_30}
\end{table*}

\ifdefined\submit
\subsection*{S1 Figure. Word clouds showing the most/least correlated words for each FA factor as obtained using Differential Language Analysis with age and gender residualized.}
{\small  Residualizing out demographics like age and gender appears to reveal other dimensions of variance like (geography, ethnicity) as illustrated by \textsc{F5} that reveals a factor highlighting language use of Indians in India with words like \texttt{india, world-cup}.}
\label{fig:wordclouds_corr_res}
\else
\begin{figure*}[htb!]
    \centering
        \includegraphics[width=\textwidth]{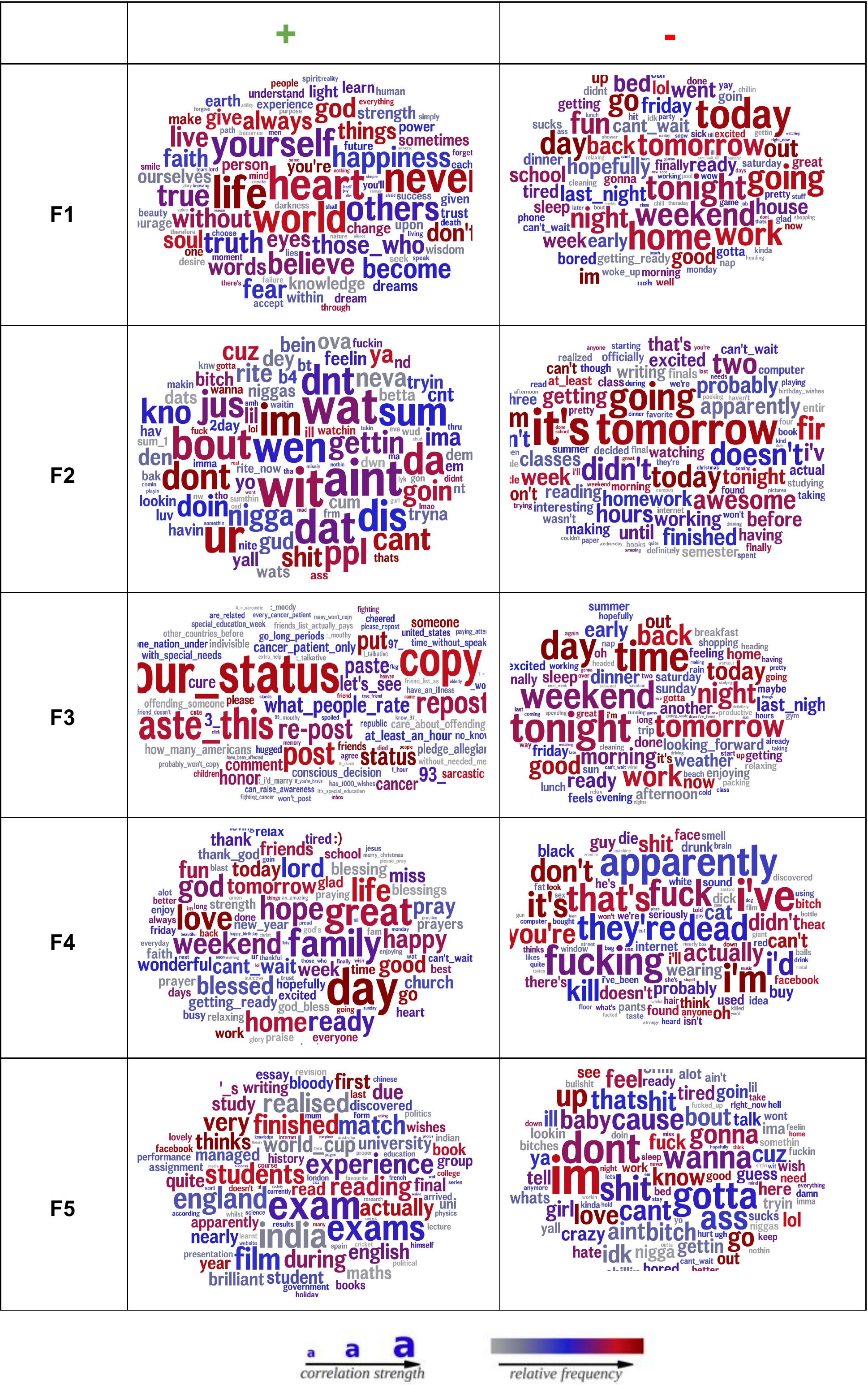}
\caption{\small Word clouds showing the most/least correlated words for each FA factor as obtained using Differential Language Analysis \textbf{with age and gender controlled by residualizing out their effect}. Residualizing out demographics like age and gender appears to reveal other dimensions of variance like (geography, ethnicity) as illustrated by \textsc{F5} that reveals a factor highlighting language use of Indians in India with words like \texttt{india, world-cup}.}
\label{fig:wordclouds_corr_res}
\end{figure*}
\fi

% Include only the SI item label in the paragraph heading. Use the \nameref{label} command to cite SI items in the text.
%\paragraph*{S1 Fig.}
%\label{S1_Fig}
%{\bf Bold the title sentence.} Add descriptive text after the title of the item (optional).

%\paragraph*{S2 Fig.}
%\label{S2_Fig}
%{\bf Lorem Ipsum.} Maecenas convallis mauris sit amet sem ultrices gravida. Etiam eget sapien nibh. Sed ac ipsum eget enim egestas ullamcorper nec euismod ligula. Curabitur fringilla pulvinar lectus consectetur pellentesque.

%\paragraph*{S1 File.}
%\label{S1_File}
%{\bf Lorem Ipsum.}  Maecenas convallis mauris sit amet sem ultrices gravida. Etiam eget sapien nibh. Sed ac ipsum eget enim egestas ullamcorper nec euismod ligula. Curabitur fringilla pulvinar lectus consectetur pellentesque.

%\paragraph*{S1 Video.}
%\label{S1_Video}
%{\bf Lorem Ipsum.}  Maecenas convallis mauris sit amet sem ultrices gravida. Etiam eget sapien nibh. Sed ac ipsum eget enim egestas ullamcorper nec euismod ligula. Curabitur fringilla pulvinar lectus consectetur pellentesque.

%\paragraph*{S1 Appendix.}
%\label{S1_Appendix}
%{\bf Lorem Ipsum.} Maecenas convallis mauris sit amet sem ultrices gravida. Etiam eget sapien nibh. Sed ac ipsum eget enim egestas ullamcorper nec euismod ligula. Curabitur fringilla pulvinar lectus consectetur pellentesque.

%\paragraph*{S1 Table.}
%\label{S1_Table}
%{\bf Lorem Ipsum.} Maecenas convallis mauris sit amet sem ultrices gravida. Etiam eget sapien nibh. Sed ac ipsum eget enim egestas ullamcorper nec euismod ligula. Curabitur fringilla pulvinar lectus consectetur pellentesque.

\section*{Acknowledgments}
 This work was supported in part by the Templeton Religion Trust, Grant TRT-0048.

\begin{comment}
\section*{Author Contributions}
Conceived and designed the experiments: VK HAS MLK SS.
Performed the experiments: VK HAS. Analyzed the data: VK HAS
Contributed reagents/materials/analysis tools: HAS MK DS SM SS.
Wrote the paper: VK HAS MLK SM DS MK LU SS
\end{comment}

\nolinenumbers

% Either type in your references using
% \begin{thebibliography}{}
% Text
% \end{thebibliography}
%
% or
%
% Compile your BiBTeX database using our plos2015.bst
% style file and paste the contents of your .bbl file
% here.
% 
\clearpage
\bibliography{paper}

\begin{thebibliography}{10}

\bibitem{allport1936trait}
Allport GW, Odbert HS.
\newblock Trait-names: A psycho-lexical study.
\newblock Psychological monographs. 1936;47(1):i.

\bibitem{cattell1946description}
Cattell RB.
\newblock Description and measurement of personality. 1946;.

\bibitem{mcadams1992five}
McAdams DP.
\newblock The five-factor model in personality: A critical appraisal.
\newblock Journal of personality. 1992;60(2):329--361.

\bibitem{john1999big}
John OP, Srivastava S.
\newblock The Big Five trait taxonomy: History, measurement, and theoretical
  perspectives.
\newblock Handbook of personality: Theory and research. 1999;2(1999):102--138.

\bibitem{costa1989neo}
Costa P, McCrae R.
\newblock Neo Five-Factor Inventory (NEO-FFI);.

\bibitem{goldberg1993structure}
Goldberg LR.
\newblock The structure of phenotypic personality traits.
\newblock American psychologist. 1993;48(1):26.

\bibitem{goldberg1963model}
Goldberg LR.
\newblock A Model of Item Ambiguity in Personality Assessment 1.
\newblock Educational and Psychological Measurement. 1963;23(3):467--492.

\bibitem{javeline1999response}
Javeline D.
\newblock Response effects in polite cultures: A test of acquiescence in
  Kazakhstan.
\newblock Public Opinion Quarterly. 1999; p. 1--28.

\bibitem{allport1927concepts}
Allport GW.
\newblock Concepts of trait and personality.
\newblock Psychological Bulletin. 1927;24(5):284.

\bibitem{iacobelli2011large}
Iacobelli F, Gill AJ, Nowson S, Oberlander J.
\newblock Large scale personality classification of bloggers.
\newblock In: Affective Computing and Intelligent Interaction. Springer; 2011.
  p. 568--577.

\bibitem{schwartz2013personality}
Schwartz HA, Eichstaedt JC, Kern ML, Dziurzynski L, Ramones SM, Agrawal M,
  et~al.
\newblock Personality, gender, and age in the language of social media: The
  open-vocabulary approach.
\newblock PloS one. 2013;8(9):e73791.

\bibitem{furnham2013dark}
Furnham A, Richards SC, Paulhus DL.
\newblock The Dark Triad of personality: A 10 year review.
\newblock Social and Personality Psychology Compass. 2013;7(3):199--216.

\bibitem{darktriad2016cikm}
Preotiuc-Pietro D, Carpenter J, Giorgi S, Ungar L.
\newblock Studying the Dark Triad of Personality through Twitter Behavior.
\newblock In: Proceedings of the 25th ACM International on Conference on
  Information and Knowledge Management. ACM; 2016. p. 761--770.

\bibitem{goldberg1981language}
Goldberg LR.
\newblock Language and individual differences: The search for universals in
  personality lexicons.
\newblock Review of personality and social psychology. 1981;2(1):141--165.

\bibitem{tausczik2010psychological}
Tausczik YR, Pennebaker JW.
\newblock The psychological meaning of words: LIWC and computerized text
  analysis methods.
\newblock Journal of language and social psychology. 2010;29(1):24--54.

\bibitem{gosling2003very}
Gosling SD, Rentfrow PJ, Swann WB.
\newblock A very brief measure of the Big-Five personality domains.
\newblock Journal of Research in personality. 2003;37(6):504--528.

\bibitem{deyoung2007between}
DeYoung CG, Quilty LC, Peterson JB.
\newblock Between facets and domains: 10 aspects of the Big Five.
\newblock Journal of personality and social psychology. 2007;93(5):880.

\bibitem{paunonen2000beyond}
Paunonen SV, Jackson DN.
\newblock What is beyond the big five? Plenty!
\newblock Journal of personality. 2000;68(5):821--835.

\bibitem{mcadams2006role}
McAdams DP.
\newblock The role of narrative in personality psychology today.
\newblock Narrative Inquiry. 2006;16(1):11--18.

\bibitem{chung2008revealing}
Chung CK, Pennebaker JW.
\newblock Revealing dimensions of thinking in open-ended self-descriptions: An
  automated meaning extraction method for natural language.
\newblock Journal of research in personality. 2008;42(1):96--132.

\bibitem{friedman2014personality}
Friedman HS, Kern ML.
\newblock Personality, well-being, and health.
\newblock Annual Review of Psychology. 2014;65:719--742.

\bibitem{roberts2007power}
Roberts BW, Kuncel NR, Shiner R, Caspi A, Goldberg LR.
\newblock The power of personality: The comparative validity of personality
  traits, socioeconomic status, and cognitive ability for predicting important
  life outcomes.
\newblock Perspectives on Psychological Science. 2007;2(4):313--345.

\bibitem{argamon2007mining}
Argamon S, Koppel M, Pennebaker JW, Schler J.
\newblock Mining the blogosphere: Age, gender and the varieties of
  self-expression.
\newblock First Monday. 2007;12(9).

\bibitem{holtgraves2011text}
Holtgraves T.
\newblock Text messaging, personality, and the social context.
\newblock Journal of research in personality. 2011;45(1):92--99.

\bibitem{sumner2011determining}
Sumner C, Byers A, Shearing M.
\newblock Determining personality traits \& privacy concerns from facebook
  activity.
\newblock Black Hat Briefings. 2011;11:197--221.

\bibitem{golbeck2011predicting}
Golbeck J, Robles C, Edmondson M, Turner K.
\newblock Predicting personality from twitter.
\newblock In: Privacy, Security, Risk and Trust (PASSAT) and 2011 IEEE Third
  Inernational Conference on Social Computing (SocialCom), 2011 IEEE Third
  International Conference on. IEEE; 2011. p. 149--156.

\bibitem{sumner2012predicting}
Sumner C, Byers A, Boochever R, Park GJ.
\newblock Predicting dark triad personality traits from twitter usage and a
  linguistic analysis of tweets.
\newblock In: Machine Learning and Applications (ICMLA), 2012 11th
  International Conference on. vol.~2. IEEE; 2012. p. 386--393.

\bibitem{plank2015personality}
Plank B, Hovy D.
\newblock Personality Traits on Twitter—or—How to Get 1,500 Personality
  Tests in a Week; 2015.

\bibitem{park2015automatic}
Park G, Schwartz HA, Eichstaedt JC, Kern ML, Kosinski M, Stillwell DJ, et~al.
\newblock Automatic personality assessment through social media language.
\newblock Journal of personality and social psychology. 2015;108(6):934.

\bibitem{liu2016language}
Liu F, Perez J, Nowson S.
\newblock A Language-independent and Compositional Model for Personality Trait
  Recognition from Short Texts.
\newblock arXiv preprint arXiv:161004345. 2016;.

\bibitem{liu2016analyzing}
Liu L, Preotiuc-Pietro D, Samani ZR, Moghaddam ME, Ungar L.
\newblock Analyzing Personality through Social Media Profile Picture Choice;
  2016.

\bibitem{kern2014online}
Kern ML, Eichstaedt JC, Schwartz HA, Dziurzynski L, Ungar LH, Stillwell DJ,
  et~al.
\newblock The online social self: An open vocabulary approach to personality.
\newblock Assessment. 2014;21(2):158--169.

\bibitem{kosinski2015facebook}
Kosinski M, Matz SC, Gosling SD, Popov V, Stillwell D.
\newblock Facebook as a research tool for the social sciences: Opportunities,
  challenges, ethical considerations, and practical guidelines.
\newblock American Psychologist. 2015;70(6):543.

\bibitem{goldberg2006international}
Goldberg LR, Johnson JA, Eber HW, Hogan R, Ashton MC, Cloninger CR, et~al.
\newblock The international personality item pool and the future of
  public-domain personality measures.
\newblock Journal of Research in personality. 2006;40(1):84--96.

\bibitem{blei2003latent}
Blei DM, Ng AY, Jordan MI.
\newblock Latent dirichlet allocation.
\newblock Journal of machine Learning research. 2003;3(Jan):993--1022.

\bibitem{mccallum2002mallet}
McCallum AK.
\newblock Mallet: A machine learning for language toolkit. 2002;.

\bibitem{landauer2006latent}
Landauer TK.
\newblock Latent semantic analysis.
\newblock Wiley Online Library; 2006.

\bibitem{diener1985satisfaction}
Diener E, Emmons RA, Larsen RJ, Griffin S.
\newblock The satisfaction with life scale.
\newblock Journal of personality assessment. 1985;49(1):71--75.

\bibitem{radloff1977ces}
Radloff LS.
\newblock The CES-D scale: A self-report depression scale for research in the
  general population.
\newblock Applied psychological measurement. 1977;1(3):385--401.

\bibitem{kuhn2010hungarian}
Kuhn HW.
\newblock The hungarian method for the assignment problem.
\newblock 50 Years of Integer Programming 1958-2008. 2010; p. 29--47.

\bibitem{mischel2004toward}
Mischel W.
\newblock Toward an integrative science of the person.
\newblock Annu Rev Psychol. 2004;55:1--22.

\end{thebibliography}

%\begin{thebibliography}{10}

%\bibitem{bib1}
%Conant GC, Wolfe KH.
%\newblock {{T}urning a hobby into a job: how duplicated genes find new
%  functions}.
%\newblock Nat Rev Genet. 2008 Dec;9(12):938--950.

%\bibitem{bib2}
%Ohno S.
%\newblock Evolution by gene duplication.
%\newblock London: George Alien \& Unwin Ltd. Berlin, Heidelberg and New York:
%  Springer-Verlag.; 1970.

%\bibitem{bib3}
%Magwire MM, Bayer F, Webster CL, Cao C, Jiggins FM.
%\newblock {{S}uccessive increases in the resistance of {D}rosophila to viral
%  infection through a transposon insertion followed by a {D}uplication}.
%\newblock PLoS Genet. 2011 Oct;7(10):e1002337.

%\end{thebibliography}

\end{document}